%% file: main.tex
\documentclass[10pt,twocolumn,letterpaper]{article}

\usepackage{iccv}             
\definecolor{iccvblue}{rgb}{0.21,0.49,0.74}
\usepackage[pagebackref,breaklinks,colorlinks,allcolors=iccvblue]{hyperref}
\usepackage{multirow}
\usepackage{xcolor}

\title{Light-A-Video: Training-free Video Relighting via Progressive Light Fusion}

\author{
Yujie Zhou\textsuperscript{1,6*},
Jiazi Bu\textsuperscript{1,6*},
Pengyang Ling\textsuperscript{2,6*},
Pan Zhang\textsuperscript{6\dag},
Tong Wu\textsuperscript{5},
Qidong Huang\textsuperscript{2,6},\\
Jinsong Li\textsuperscript{3,6},
Xiaoyi Dong\textsuperscript{6},
Yuhang Zang\textsuperscript{6}, 
Yuhang Cao\textsuperscript{6},
Anyi Rao\textsuperscript{4},
Jiaqi Wang\textsuperscript{6},
Li Niu\textsuperscript{1\dag}\\
\normalsize \textsuperscript{1}Shanghai Jiao Tong University \quad 
\textsuperscript{2}University of Science and Technology of China \quad 
\textsuperscript{3}The Chinese University of Hong Kong \\
\normalsize \textsuperscript{4}Hong Kong University of Science and Technology \quad 
\textsuperscript{5}Stanford University \quad 
\textsuperscript{6}Shanghai AI Laboratory \\
}

\begin{document}

\twocolumn[{
\maketitle
\vspace{-8mm}
\begin{center}
    \centering
    \captionsetup{type=figure}
    \includegraphics[width=\textwidth]{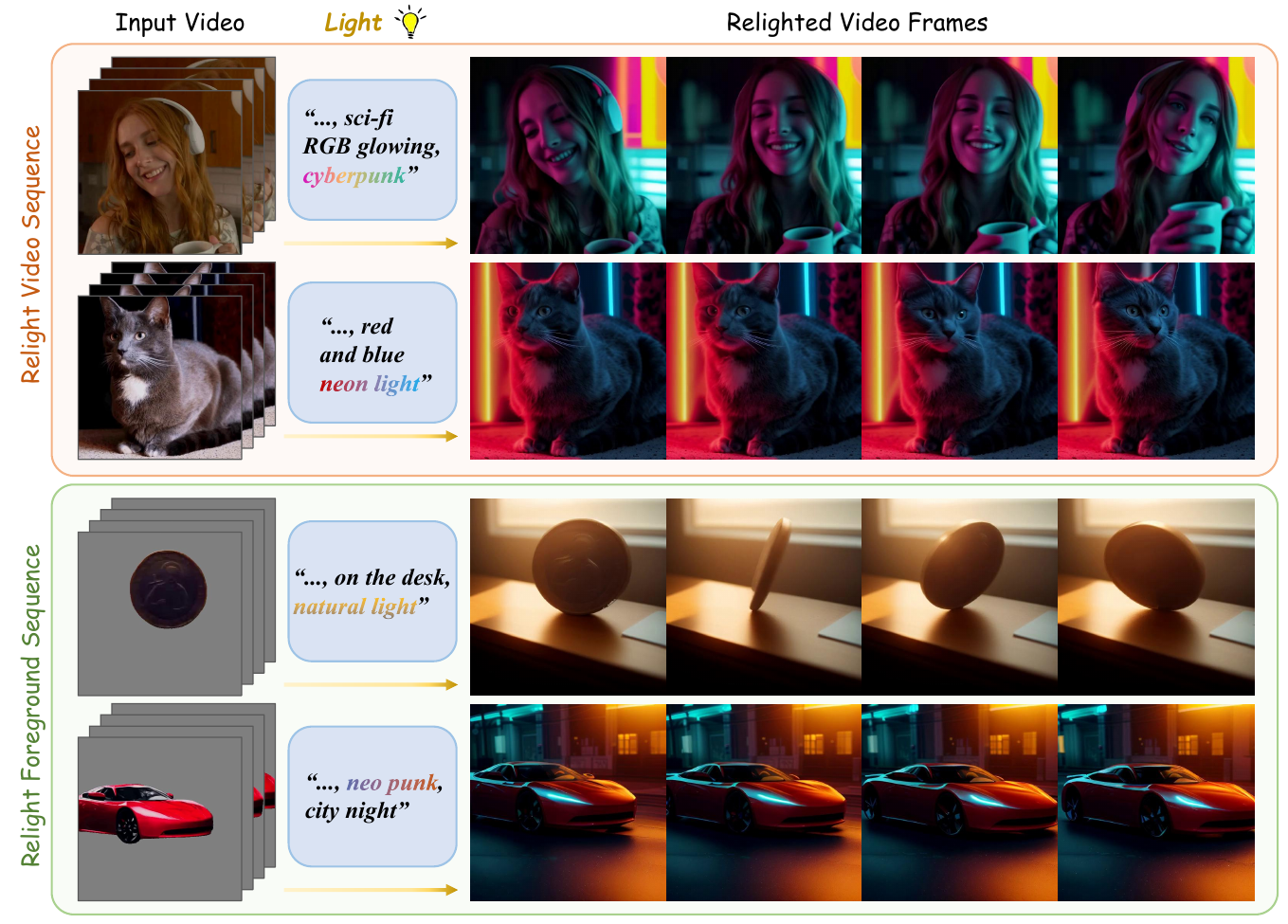}
    \vspace{-5mm}
    \captionof{figure}{{\bf  Training-free video relighting.} 
    Equipped with an image relighting model (\textit{e.g.},
    IC-Light~\citep{zhang2025scaling}) and a video diffusion model (\textit{e.g.},
    CogVideoX~\citep{yang2024cogvideox} and AnimateDiff~\citep{guo2023animatediff}), 
    Light-A-Video enables training-free video relighting 
    for given video sequences or foreground sequences.
    }
    \label{fig:teaser}
\end{center}
}]

\input{sec/0_abstract}    
\input{sec/1_intro}
\input{sec/2_related}
\input{sec/3_preliminary}
\input{sec/4_method}
\input{sec/5_experiment}
\input{sec/6_conclusion}

{
    \small
    \bibliographystyle{ieeenat_fullname}
    \bibliography{main}
}
\input{sec/supp}

\end{document}

%% file: sec/0_abstract.tex
\begin{abstract}
Recent advancements in image relighting models, 
driven by large-scale datasets and pre-trained diffusion models, 
have enabled the imposition of consistent lighting. 
However, video relighting still lags, 
primarily due to the excessive training costs and the
scarcity of diverse, high-quality video relighting datasets.
A simple application of image relighting models on a frame-by-frame basis leads to several issues: 
lighting source inconsistency and relighted appearance inconsistency,
resulting in flickers in the generated videos.
In this work, we propose \textbf{Light-A-Video}, 
a training-free approach to achieve temporally smooth video relighting.
Adapted from image relighting models, 
Light-A-Video introduces two key techniques to enhance lighting consistency.
First, we design a Consistent Light Attention \textbf{(CLA)} module, 
which enhances cross-frame interactions within the self-attention layers of the image relight model 
to stabilize the generation of the background lighting source.
Second, leveraging the physical principle of light transport independence, 
we apply linear blending between the source video’s appearance
and the relighted appearance, 
using a Progressive Light Fusion \textbf{(PLF)} strategy to ensure
smooth temporal transitions in illumination.
Experiments show that Light-A-Video improves the temporal consistency of relighted video
while maintaining the relighted image quality, 
ensuring coherent lighting transitions across frames.
Project page: \url{https://bujiazi.github.io/light-a-video.github.io/}.
\vspace{-1.5em}
\end{abstract}

%% file: sec/1_intro.tex
\section{Introduction}
\label{sec:intro}
Illumination plays a crucial role in shaping our perception of visual content,
impacting both its aesthetic quality and human interpretation of scenes. 
Relighting tasks~\citep{sun2019single, nestmeyer2020learning, pandey2021total, zhou2019deep, sengupta2021light, hou2021towards, wang2023sunstage, zhou2023relightable, kocsis2024lightit}, which focus on adjusting lighting conditions in 2D and 3D visual content,
have long been a key area of research in computer graphics due to their broad practical applications,
such as film production, gaming, and virtual environments.
Traditional image relighting methods, which rely on physical illumination models, are typically confined to controlled laboratory settings and struggle to generalize to complex, unconstrained real-world lighting and material estimation.

In order to address these limitations, data-driven approaches~\citep{deng2025flashtex, ren2024relightful, kim2024switchlight, zhang2024lumisculpt, zeng2024dilightnet, jin2024neural} have emerged, 
leveraging large-scale, diverse relighting datasets combined with pre-trained diffusion models. 
As the state-of-the-art image relighting model, IC-Light~\citep{zhang2025scaling} modifies only the illumination of an image 
while maintaining its albedo unchanged.
Based on the physical principle of light transport independence, 
IC-Light allows for controllable and stable illumination editing, 
such as adjusting lighting effects and simulating complex lighting scenarios.
However, video relighting is significantly more challenging
as it needs to maintain temporal consistency across frames. 
The scarcity of video lighting datasets and the high training costs further complicate the task. 
Thus, existing video relighting methods~\citep{zhang2024lumisculpt}
struggle to deliver consistently high-quality results or
are limited to specific domains, such as portraits.

\begin{figure}
\centering
\includegraphics[width=\linewidth]{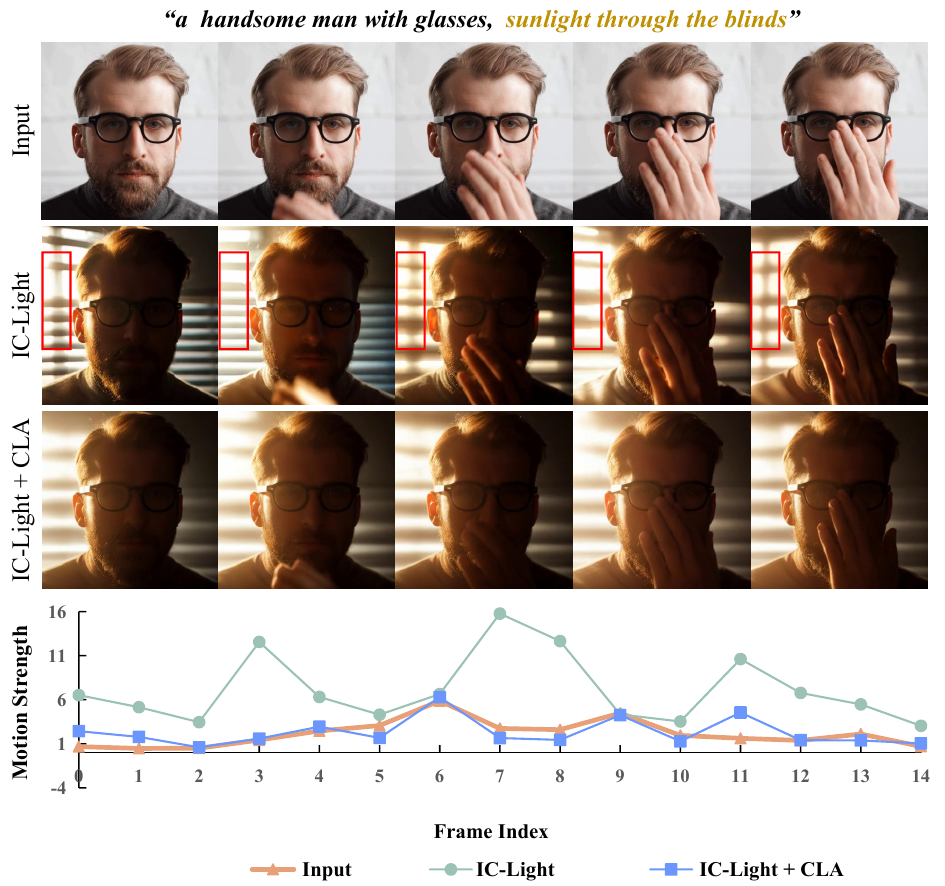}
\vspace{-2em}
\caption{{\bf Relighted frames of vanilla IC-Light and ``IC-Light + CLA" }. 
The line chart depicts the average optical flow intensity between adjacent frames. Since IC-Light performs image relighting based on each independent frame, its results show a noticeable jitter between frames, especially in the generated background lighting. Conversely, the proposed CLA facilitates consistent lighting generation by forcing interaction between frames.
}
\label{fig:cla}
\vspace{-1.5em}
\end{figure}
In this work, we propose a training-free approach for video relighting,
named \textbf{Light-A-Video}, which enables the generation of smooth, high-quality relighted videos
without any additional training or optimization. 
As shown in Fig.~\ref{fig:teaser}, given a text prompt that provides a general description of the video and 
specified illumination conditions, 
our Light-A-Video pipeline can relight the input video in a zero-shot manner,
fully leveraging the relighting capabilities of image-based models and the motion priors of the video diffusion model.
To achieve this goal, we initially apply an image-relighting model to video-relighting tasks on a frame-by-frame basis,
and observe that the generated lighting source is unstable across the video frames.
This instability leads to inconsistencies in the relighting of the objects' appearances and significant flickering across frames.
To stabilize the generated lighting source and ensure consistent results, we design a
Consistent Light Attention \textbf{(CLA)} module within the self-attention layers of the image relighting model. 
As shown in Fig.~\ref{fig:cla},
by incorporating additional temporally averaged features into the attention computation,
CLA facilitates cross-frame interactions, producing a structurally stable lighting source.
To further enhance the appearance stability across frames,
we utilize the motion priors of the video diffusion model with a novel Progressive Light Fusion \textbf{(PLF)} strategy.
Adhering to the physical principles of light transport,
PLF progressively employs linear blending to integrate relighted appearances from the CLA 
into each original denoising target, 
which gradually guides the video denoising process toward the desired relighting direction.
Finally, Light-A-Video serves as a complete end-to-end pipeline, effectively achieving smooth and consistent video relighting.
As a training-free approach, Light-A-Video is not restricted to specific video diffusion models,
making it highly compatible with a range of popular video generation backbones, 
including UNet-based and DiT-based models such as AnimateDiff~\citep{guo2023animatediff} and
CogVideoX~\citep{yang2024cogvideox}.
Our contributions are summarized as follows:
\begin{itemize}

\item We propose Light-A-Video, a novel training-free video relighting framework that
generalizes the capabilities of image relighting models to the video domain, enabling flexible and temporally consistent video relighting.


\item We introduce two key designs: a consistent light attention module to enhance the stability of lighting sources across frames,
and a progressive light fusion strategy gradually injects lighting information to facilitate temporal consistency in video appearance.

\item Extensive experiments under various scenarios demonstrate the effectiveness and versatility of the proposed method, which not only supports relighting the entire video sequences
but also enables relighting for given foreground sequences.

\end{itemize}
\vspace{-0.5em}

%% file: sec/2_related.tex
\section{Related Work}
\label{sec:related}

\noindent{\textbf{Video Diffusion Models.}}
Video diffusion models~\citep{blattmann2023align, chen2023videocrafter1, chen2024videocrafter2, guo2023animatediff, wang2023modelscope, wang2023lavie, hong2022cogvideo, yang2024cogvideox, blattmann2023stable, zhang2024show, bu2024broadway} aim to synthesize temporally consistent image frames based on provided conditions, such as a text prompt or an image prompt. In the realm of text-to-video (T2V) generation, the majority of methods~\citep{wang2023modelscope, guo2023animatediff, chen2023videocrafter1, chen2024videocrafter2, blattmann2023align, zhang2024show} train additional motion modeling modules from existing text-to-image architectures to model the correlation between video frames, while others~\citep{hong2022cogvideo, yang2024cogvideox} train from scratch to learn video priors. For image-to-video (I2V) tasks that enhance still images with reasonable motions, a line of research~\citep {xing2025dynamicrafter, chen2025livephoto} proposes novel frameworks dedicated to image animation. Some approaches~\citep{guo2023i2v, zhang2024pia, guo2025sparsectrl} serve as plug-to-play adapters. Stable Video Diffusion~\citep{blattmann2023stable} fine-tune pre-trained T2V models for I2V generation, achieving impressive performance. Numerous works~\cite {niu2025mofa, ma2024follow,ling2024motionclone} focus on controllable generation, providing more controllability for users. Video diffusion models, due to their inherent video priors, are capable of synthesizing smooth and consistent video frames that are both content-rich and temporally harmonious.

\noindent{\textbf{Learning-based Illumination Control.}}
Over the past few years, a variety of lighting control techniques~\citep{sun2019single, nestmeyer2020learning, pandey2021total} for 2D and 3D visual content based on deep neural networks have been proposed, especially in the field of portrait lighting~\citep{shu2017portrait, barron2014shape, sengupta2018sfsnet, shih2014style, kim2024switchlight}, along with a range of baselines~\citep{zhou2019deep, sengupta2021light, hou2021towards, wang2023sunstage, zhou2023relightable} aimed at improving the effectiveness, accuracy, and theoretical foundation of illumination modeling. Recently, owing to the rapid development of diffusion-based generative models, a number of lighting control methods~\citep{ren2024relightful, zeng2024dilightnet, deng2025flashtex, jin2024neural} utilizing diffusion models have also been introduced. Relightful Harmonization~\citep{ren2024relightful} focuses on harmonizing sophisticated lighting effects for the foreground portrait conditioning on a given background image.
SwitchLight~\citep{kim2024switchlight} suggests training a physically co-designed framework for human portrait relighting. IC-Light~\citep{zhang2025scaling} is a state-of-the-art approach for image relighting. LumiSculpt~\citep{zhang2024lumisculpt} enables consistent lighting control in T2V generation models for the first time. However, in the domain of video lighting, the aforementioned approaches fail to simultaneously ensure precise lighting control and exceptional visual quality. This work incorporates a pre-trained image lighting control model into the denoising process of a T2V model through progressive guidance, leveraging the latter's video priors to facilitate the smooth transfer of image lighting control knowledge, thereby enabling accurate and harmonized control of video lighting.

\noindent{\textbf{Video Editing with Diffusion Models.}}
In recent years, diffusion-based video editing has undergone significant advancements. 
Some researches~\citep{liu2024video, wang2023zero, wu2023tune, ma2024follow, mokady2023null} 
adopt pre-trained text-to-image (T2I) backbones for video editing. Another line of approaches~\citep{yang2024fresco, yang2023rerender, cong2023flatten, hu2023videocontrolnet} leverages pre-trained optical flow models to enhance the temporal consistency of output video. Numerous studies~\citep{qi2023fatezero, geyer2023tokenflow, kara2024rave} have concentrated on exploring zero-shot video editing approaches. COVE~\citep{wang2024cove} leverages the inherent diffusion feature correspondence proposed by DIFT~\citep{tang2023emergent} to achieve consistent video editing. SDEdit~\citep{meng2021sdedit} utilizes the intrinsic capability of diffusion models to refine details based on a given layout, enabling efficient editing for both image and video. Despite the remarkable performance of existing video editing techniques in various settings, there remains a lack of approaches specifically designed for controlling the lighting of videos. 

%% file: sec/3_preliminary.tex
\vspace{-0.5em}
\section{Preliminary}
\label{sec:preliminary}

\begin{figure*}
\centering
\includegraphics[width=\linewidth]{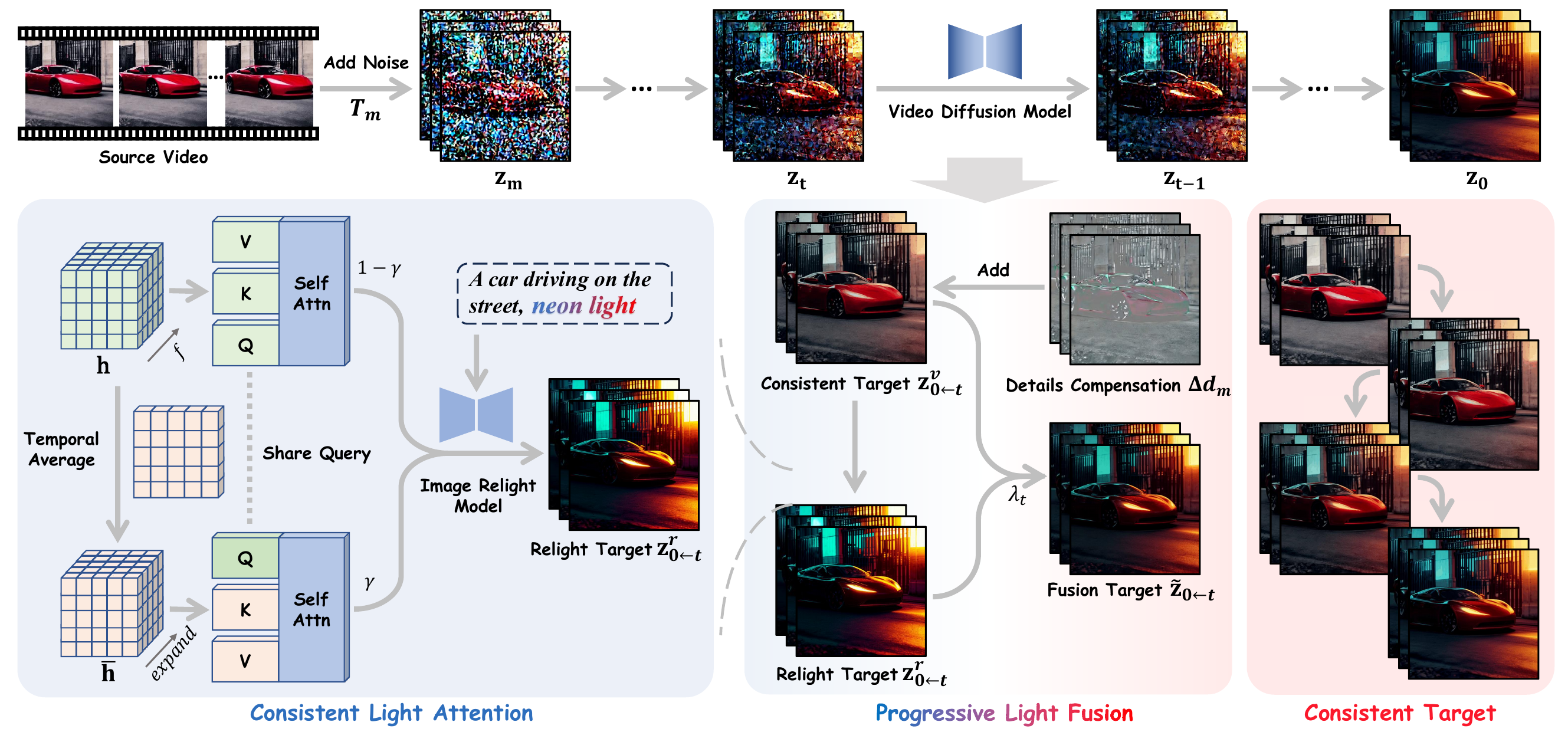}
\vspace{-1.0em}
\caption{{\bf The pipeline of Light-A-Video}. A source video is first noised and processed through the VDM for denoising across $T_m$
steps. At each step, the predicted noise-free component with details compensation serves as the Consistent Target $\mathbf{z}^{v}_{0 \gets t}$, 
inherently representing the VDM's denoising direction. 
Consistent Light Attention infuses $\mathbf{z}^{v}_{0 \gets t}$ with unique lighting information,
transforming it into the Relight Target $\mathbf{z}^{r}_{0 \gets t}$.
The Progressive Light Fusion strategy then merges two targets to form the Fusion Target $\tilde{\mathbf{z}}_{0 \gets t}$, 
which provides a refined direction for the current step.The bottom-right part illustrates the iterative evolution of $\mathbf{z}^{v}_{0 \gets t}$.}
\label{fig:pipe}
\vspace{-1.2em}
\end{figure*}

\subsection{Diffusion Model}
\label{sec:dm}
Given an image $\mathbf{x}_0$ that follows the real-world data distribution, we first 
encode $\mathbf{x}_0$ into latent space $\mathbf{z}_0 = \mathcal{E}(\mathbf{x}_0)$
using a pretrained autoencoder $\{ {\mathcal{E}(\cdot)},{\mathcal{D}(\cdot)} \}$.
The forward diffusion process is a $T$ steps Markov chain~\cite{ho2020denoising},
corresponding to the iterative introduction of Gaussian noise $\epsilon$, which can be expressed as:
\begin{equation}
    \mathbf{z}_t =  \sqrt{1-\beta _t}\mathbf{z}_{t-1} + \sqrt{\beta _t} \epsilon, 
\end{equation}
where $\beta _t \in (0,1)$ determines the amount of Gaussian noise introduced at time step $t$. Mathematically, the above cumulative noise adding has the following closed-form:
\begin{equation}
    \mathbf{z}_t = \sqrt{\bar{\alpha}_t}  \mathbf{z}_{0} + \sqrt{1-\bar{\alpha}_t}\epsilon, 
\end{equation}
where $\bar{\alpha}_t =  {\textstyle \prod_{1}^{t}} (1-\beta _t)$. 

For numerical stability, $\mathbf{v}$-prediction~\cite{salimans2022progressive} approach is employed, 
where the diffusion model outputs a predicted velocity $\mathbf{v}$ to represent the denoising direction.
Defined as:
\begin{equation}\label{eq: v_pred}
    \mathbf{v}=\sqrt{\bar{\alpha}_t} \epsilon-\sqrt{1-\bar{\alpha}_t} \mathbf{z}_0.
\end{equation}
During inference, the noise-free component $\hat{\mathbf{z}}_{0 \gets t}$ can be recovered from the model's output $\mathbf{v}_t$ as follows:
\begin{equation}\label{eq: z0_pred}
    \hat{\mathbf{z}}_{0 \gets t}=\sqrt{\bar{\alpha}_t} \mathbf{z}_t-\sqrt{1-\bar{\alpha}_t} \mathbf{v}_t.
\end{equation}
$ \hat{\mathbf{z}}_{0 \gets t}$ represents the denoising target at time step $t$.

\subsection{ Light Transport}
\label{sec:lightTrans}
Light transport theory~\cite{debevec2000acquiring,zhang2025scaling} demonstrates that arbitrary image appearance
$\mathbf{I}_L$ can be decomposed by the product of a light transport matrix $\mathbf{T}$  and environment illumination $L$, which can be expressed as:
\begin{equation}\label{eq:decomposition}
    \mathbf{I}_L = \mathbf{T}L,
\end{equation}
where $\mathbf{T}$ is a single matrix for linear light transform~\cite{debevec2000acquiring} and 
$L$ denotes variable environment illumination.
Given the linearity of $\mathbf{T}$, the merging between environment illumination $L$ 
is equal to the fusion of image appearance $\mathbf{I}_L$, i.e., 
\begin{equation}\label{eq:light_fusion}
    \mathbf{I}_{L_1 + L_2} =  \mathbf{T}(L_1 +L_2) = \mathbf{I}_{L_1} + \mathbf{I}_{L_2}.
\end{equation}
Such characteristic suggests the feasibility of lighting control by indirectly constraining image appearance, 
i.e., the consistent image light constraint in IC-Light~\citep{zhang2025scaling}.

\label{sec:iclight}

%% file: sec/4_method.tex
\section{Light-A-Video}
\label{sec:method}


Section~\ref{sec:PF} defines the objectives of the video relighting task. Section~\ref{sec:CLA} reveals that per-frame image relighting for video sequences suffers from lighting source inconsistency
and accordingly proposes the Consistent Lighting Attention \textbf{(CLA)} module for enhanced stability in generated lighting source across frames. Section~\ref{sec:PLF} represents the Progressive Light Fusion \textbf{(PLF)} strategy for temporally consistent video
appearance generation.
\subsection{Problem Formulation}
\label{sec:PF}
Given a source video and a lighting condition $c$, 
the objective of video relighting is to
render the source video into the relighted video that maintains the motion in the source video while aligning the lighting condition $c$.
Unlike image relighting that solely concentrates on appearance, video relighting raises extra challenges in
maintaining temporal consistency and motion preservation, necessitating high-quality visual coherence across frames.
\subsection{Consistent Light Attention}
\label{sec:CLA}
Given the achievement in image relighting model~\cite{zhang2025scaling}, a straightforward approach for video relighting is to
directly perform frame-by-frame image relighting under the same lighting condition. 
However, as illustrated in Fig.~\ref{fig:cla}, this naive method fails to maintain appearance coherence across frames,
resulting in frequent flickering of the generated light source
and inconsistent temporal illumination.

To improve inter-frame information integration and generate a stable light source,
we propose a Consistent Light Attention \textbf{(CLA)} module.
Specifically, for each self-attention layer in the IC-Light model, 
a video feature map $\mathbf{h} \in \mathbb{R}^{(b \times f) \times (h \times w) \times d}$ serves as 
the input, where $b$ is the batch size and $f$ is the number of video frames, $h$ and $w$ denote the height and width of the feature map, 
with $h \times w$ representing the number of tokens for attention computation.
With linearly projections, $\mathbf{h}$ is projected into query, 
key and value features $Q, K, V \in \mathbb{R}^{ (b \times f) \times (h \times w) \times d}$.
The attention computation is defined as follows:
\begin{equation}
\operatorname{Self-Attn}(Q, K, V)=\operatorname{Softmax}\left(\frac{Q K^T}{\sqrt{d}}\right) V.
\end{equation}
Note that the naive method treats the frame dimension as the batch size, 
performing self-attention frame by frame with the image relighting model, 
which results in each frame attending only to its features.
For the CLA module, as shown in Fig.~\ref{fig:pipe},
a dual-stream attention fusion strategy is applied.
Given the input feature $\mathbf{h}$, the original stream directly feeds the feature map into the attention module to 
compute frame-by-frame attention, resulting in the output $\mathbf{h}^{\prime}_1$.
The average stream first reshapes $\mathbf{h}$ into $ \mathbb{R}^{b \times f \times (h \times w) \times d}$, averages
it along the temporal dimension, then expands it $f$ times to obtain $\mathbf{\bar{h}}$.
Specifically, the average stream mitigates high-frequency temporal fluctuations,
thereby facilitating the generation of a stable background light source across frames.
Meanwhile, the original stream
retains the original high-frequency details, 
thereby compensating for the detail loss incurred by the averaging process.
Then,
$\mathbf{\bar{h}}$ is input into the self-attention module and the output is $\mathbf{\bar{h}}^{\prime}_2$.
The final output $\mathbf{h}_o^{\prime}$ of the CLA module is a weighted average between two streams, with the trade-off parameter $\gamma$,
\begin{equation}\label{eq:attn_fusion}
\mathbf{h}_o^{\prime} = (1-\gamma) \mathbf{h}^{\prime}_1 + \gamma \mathbf{\bar{h}}^{\prime}_2.
\end{equation}
With the help of CLA, 
the result can capture global context across the entire video
and generate a more stable lighting source, as shown in Fig.~\ref{fig:cla}.

\subsection{Progressive Light Fusion}
\label{sec:PLF}

CLA module improves cross-frame consistency but lacks pixel-level constraints, 
leading to inconsistencies in appearance details.
To address this, we leverage motion priors in the Video Diffusion Model (VDM), 
which are trained on large-scale video datasets and use a temporal attention module
to ensure consistent motion and lighting changes.
The novelty of our Light-A-Video lies in 
progressively injecting the relighting results
as guidance into the denoising process.

In the pipeline as shown in Fig~\ref{fig:pipe},
a source video is first encoded into latent space, 
and then add $T_m$ step noise to acquire the noisy latent $\mathbf{z}_m$.
At each denoising step $t$,
the noise-free component $\hat{\mathbf{z}}_{0 \gets t}$ 
in Eq.~\ref{eq: z0_pred} is predicted, which serves as the denoising target for the current step.
Prior work demonstrated the potential of applying tailored manipulation in denoising targets for guided generation, 
with significant achievements observed in high-resolution image synthesis~\cite{kim2024diffusehigh} 
and text-based image editing~\cite{rout2024rfinversion}.

\begin{figure}
\centering
\includegraphics[width=0.9\linewidth]{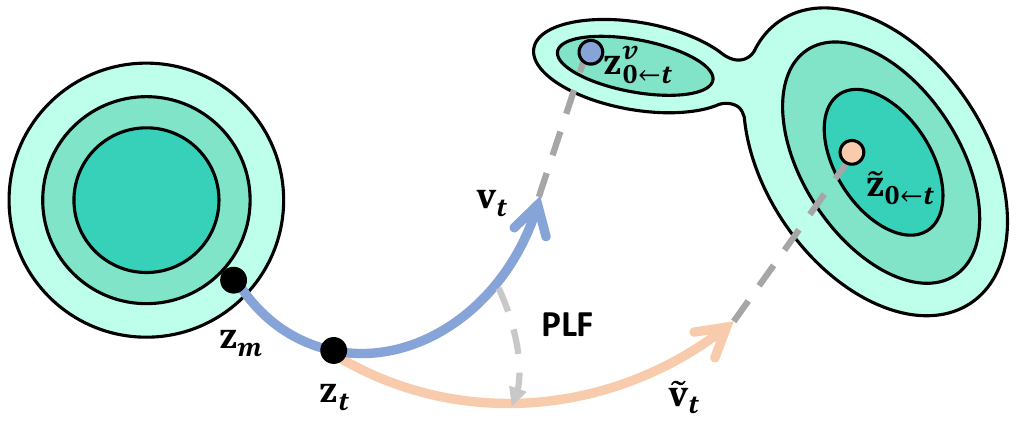}
\vspace{-0.5em}
\caption{{\bf Visualization of the PLF strategy}. 
During the denoising process of the VDM, the PLF strategy progressively
replaces the original Consistent Target $\mathbf{z}^{v}_{0 \gets t}$ with the Fusion Target $\tilde{\mathbf{z}}_{0 \gets t}$,
guiding the denoising direction from  $ \mathbf{v}_t$ to $\tilde{\mathbf{v}}_t$.
}
\label{fig:plf}
\vspace{-0.5em}
\end{figure}

\begin{figure}
\centering
\includegraphics[width=\linewidth]{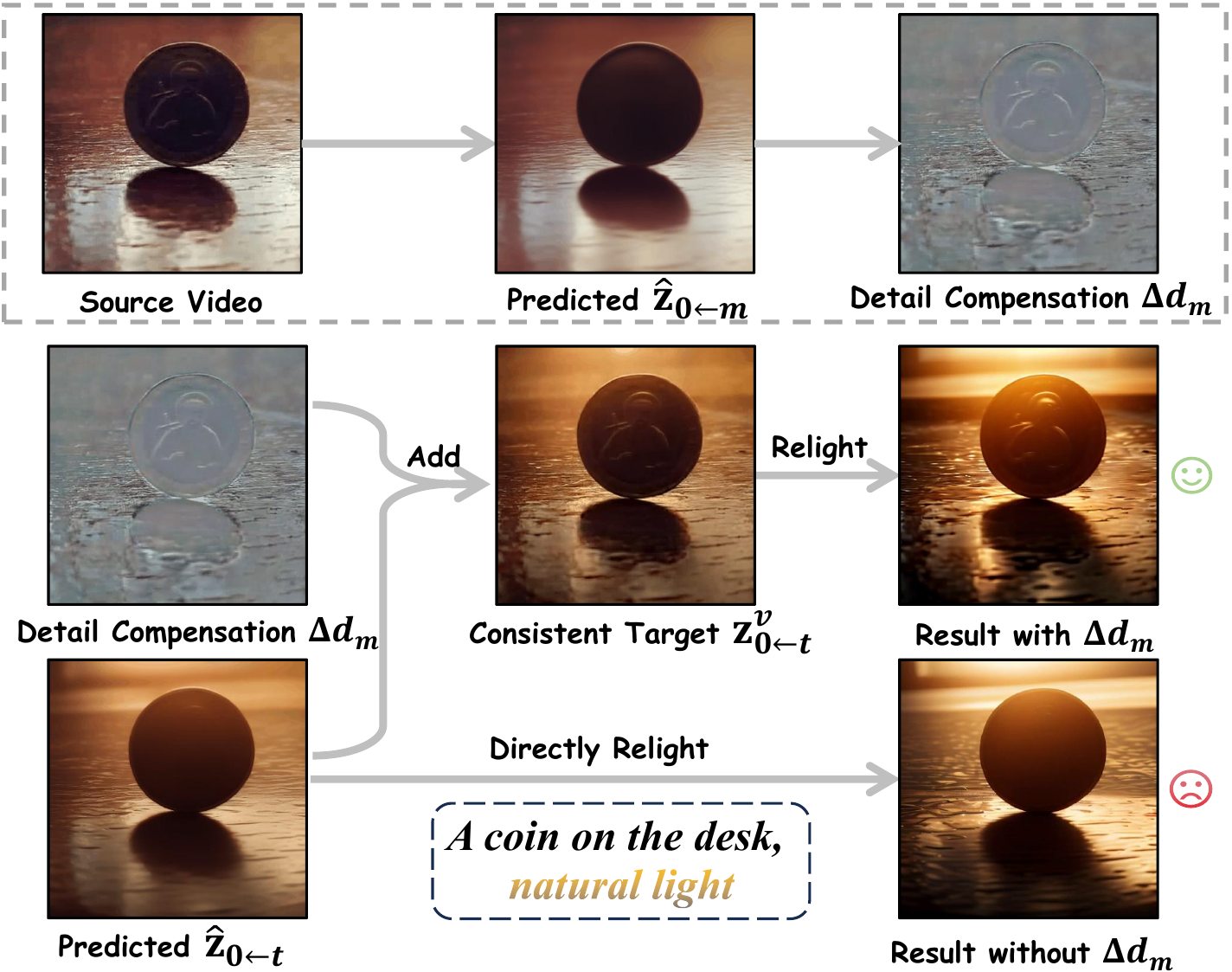}
\vspace{-1.9em}
\caption{{\bf Visualization of the detail compensation}. 
$\Delta d_m$ records the difference between $\hat{\mathbf{z}}_{0 \gets m}$ and the source video in the first denoising step, which is used
as a detail compensation component for detail preservation in the consistent target.
}
\label{fig:detail}
\vspace{-1.5em}
\end{figure}

Driven by the motion priors in the VDM, the denoising process encourage 
$\hat{\mathbf{z}}_{0 \gets t}$ to be temporally consistent.
Thus, we define this target as the video \textbf{Consistent Target $\mathbf{z}^{v}_{0 \gets t}$} with environment illumination $L^{v}_t$.
However, discrepancies still exist between the predicted $\hat{\mathbf{z}}_{0 \gets t}$ and the original video,
resulting in detail loss in the relighted video.
To address this issue, as shown in Fig.~\ref{fig:detail},
details compensation $\Delta d_m$ is incorporated into the $\mathbf{z}^{v}_{0 \gets t}$ at each step.
Then, $\mathbf{z}^{v}_{0 \gets t}$ is sent into the CLA module to
obtain the relighted latent, which serves 
as the \textbf{Relight Target $\mathbf{z}^{r}_{0 \gets t}$} with the illumination $L^{r}_t$ for the $t$-th denoising step.
Aligning with the light transport theory in Section~\ref{sec:lightTrans},
a pre-trained VAE $\{ {\mathcal{E}(\cdot)}, {\mathcal{D}(\cdot)} \}$ is used to
decode the two targets into pixel level,
yielding the image appearances 
$\mathbf{I}^{v}_{t} = \mathcal{D}(\mathbf{z}^{v}_{0 \gets t})$ 
and $ \mathbf{I}^{r}_{t} = \mathcal{D}(\mathbf{z}^{r}_{0 \gets t})$, respectively.
Refer to Eq.~\ref{eq:light_fusion},
the fusing appearance $\mathbf{I}^{f}_{t}$ can be formulated as:
\begin{equation} \label{eq:target_fusion}
   \mathbf{I}^{f}_{t} =  \mathbf{T}(L^{v}_t +L^{r}_t) = \mathbf{I}^{v}_{t} + \mathbf{I}^{r}_{t}.
\end{equation}
It is observed that directly using encoded latent $\mathcal{E}(\mathbf{I}^{f}_{t})$
as the new target at each step results in suboptimal performance.
This is attributed to the excessively large gap between the two targets, 
which exceeds the refinement capability of the VDM and 
consequently causes visible temporal lighting jitter.
To mitigate this gap, a progressive lighting fusion strategy is proposed.
Specifically, a fusion weight $\lambda_t$ is introduced, which 
decreases as denoising progresses, thereby 
gradually reducing the influence of the relight target.
The progressive light fusion appearance is defined as $\mathbf{I}^{p}_{t}$, i.e.,
\begin{equation} \label{eq:pro_target_fusion}
   \mathbf{I}^{p}_{t} = \mathbf{I}^{v}_{t} + \lambda_t (\mathbf{I}^{r}_{t}-\mathbf{I}^{v}_{t}).
\end{equation}
The encoded latent $ \tilde{\mathbf{z}}_{0 \gets t} = \mathcal{E}(\mathbf{I}^{p}_{t})$ is utilized as the \textbf{Fusion Target} for step $t$,
replacing the original $\mathbf{z}^{v}_{0 \gets t}$.
Based on the fusion target, the less noisy latent $\mathbf{z}_{t-1}$ can be computed with DDIM scheduler with $v$-prediction:
\begin{equation}
    a_t=\sqrt{\frac{1-\bar{\alpha}_{t-1}}{1-\bar{\alpha}_t}},\\
    b_t=\sqrt{\bar{\alpha}_{t-1}}-\sqrt{\bar{\alpha}_t}  a_t,
\end{equation}
\begin{equation}
\mathbf{z}_{t-1}=a_t \mathbf{z}_t+b_t  \tilde{\mathbf{z}}_{0 \gets t}.
\end{equation}
From Eq.~\ref{eq: z0_pred}, the fusion target $ \tilde{\mathbf{z}}_{0 \gets t} $ determines a new denoising direction,
denoted as $\tilde{\mathbf{v}}_t$,
\begin{equation}
\tilde{\mathbf{v}}_t=\frac{\sqrt{\bar{\alpha}_t} \mathbf{z}_t-\tilde{\mathbf{z}}_{0 \gets t}}{\sqrt{1-\bar{\alpha}_t}},
\end{equation}
which means PLF essentially refines $\mathbf{v}_t$ iteratively and
guides the denoising process towards the relighting direction, as shown in Fig~\ref{fig:plf}.
Other schedulers capable of modeling the denoising direction, such as Euler Scheduler~\cite{karras2022elucidating} and Rectified Flow~\cite{lipman2022flow}, are also applicable.
As the denoising progresses, smooth and consistent illumination injection is achieved, ensuring coherent video relighting.

%% file: sec/5_experiment.tex
\vspace{-0.5em}
\section{Experiments}
\label{sec:experiment}

\begin{figure*}[htp]
\centering
\includegraphics[width=1.0\linewidth]{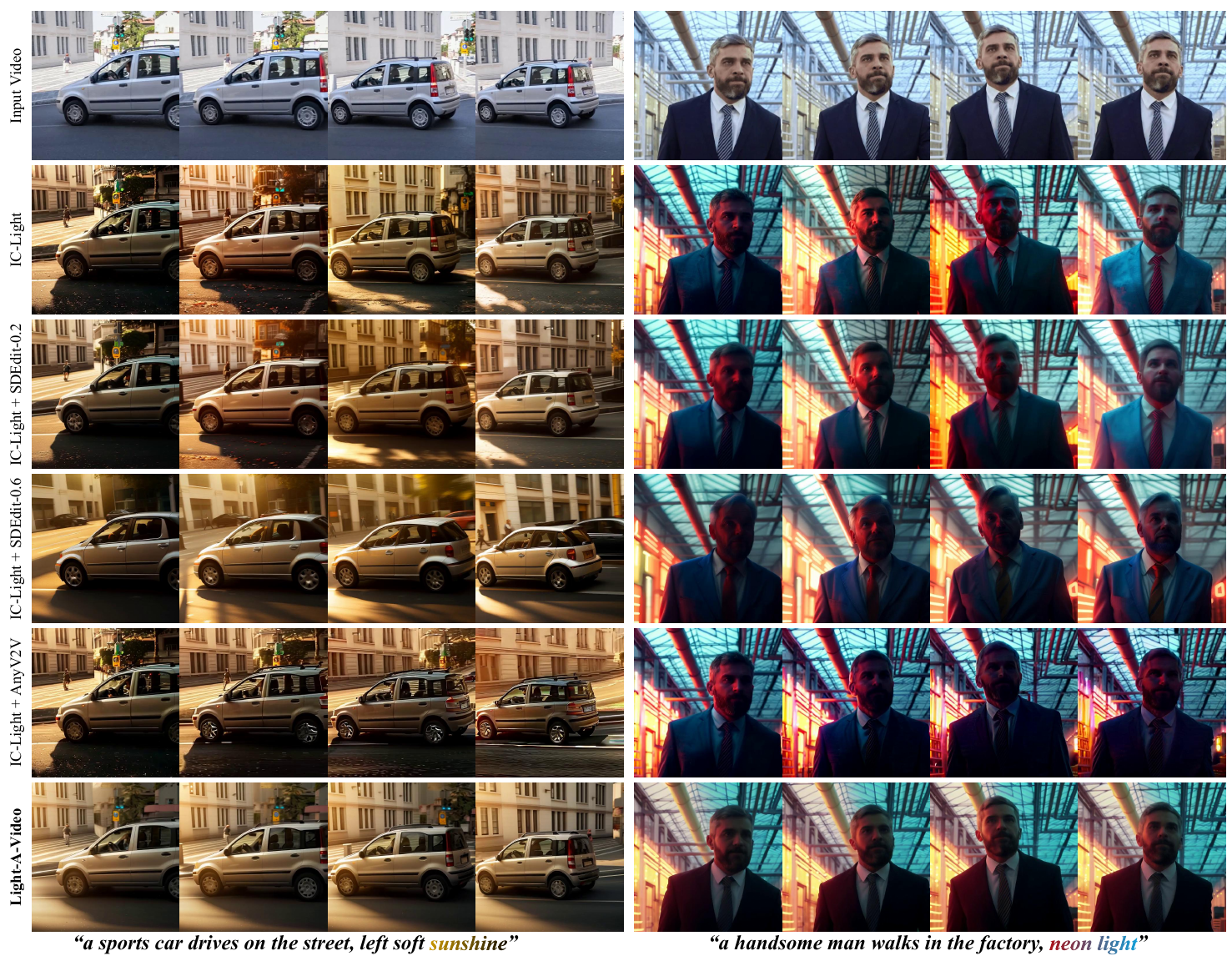}
\vspace{-2em}
\caption{\textbf{Qualitative comparison of baseline methods}.
Given a source video and guidance text prompt, 
Light-A-Video achieves high temporal consistency and fidelity to the light condition, 
outperforming other methods in avoiding flickering, jitter, and identity shifts.
VDM used: AnimateDiff (Left), CogVideoX (Right).
}
\vspace{-0.5em}
\label{fig:comparison}
\end{figure*}

\input{./tabs/tab_1}

\subsection{Experimental Details}
\label{sec:details}

\noindent \textbf{Baselines.} 
Given the lack of established video relighting methods, we adopt the state-of-the-art image 
relighting technique to perform frame-by-frame relighting on videos as a baseline. 
To verify the temporal smoothing effect of illumination using a VDM,
we construct two comparative methods by applying SDEdit~\citep{meng2021sdedit} to the per-frame results of IC-Light~\citep{zhang2025scaling}.
These two methods are named IC-Light + SDEdit-0.2 and IC-Light + SDEdit-0.6, 
corresponding to noise levels of 20\% and 60\%.
Finally, IC-Light + AnyV2V~\citep{ku2024anyv2v} serves as another baseline. 
Specifically, IC-Light relights the first frame, and AnyV2V propagates the appearance information
from the first frame to all subsequent frames, preserving the content of the source video.

\noindent \textbf{Evaluation metrics.} Three widely adopted metrics are reported for quantitative evaluation.
First, the temporal consistency of the generated video is assessed using the
average CLIP~\citep{radford2021learning} score across consecutive video frames. 
Second, the optical flow for each baseline video is estimated using RAFT~\citep{teed2020raft},
and the motion preservation score of each method is assessed by calculating the optical flow discrepancy with the source video.
Third, to evaluate the quality of relighted image, a video test dataset is collected.
The FID~\cite{Seitzer2020FID} score is then calculated between the results of each method
and the frame-by-frame IC-Light results, serving as the metric for relight quality evaluation.
Finally, 52 volunteers are invited to conduct a user study across three aspects: 
\textbf{L}ighting \textbf{P}rompt \textbf{A}lignment (alignment between video content and lighting prompt), 
\textbf{V}ideo \textbf{S}moothness (temporal consistency of the relighted video), and \textbf{I}D-\textbf{P}reservation 
(consistency of the object’s identity and albedo
before and after relighting). 
The volunteers rank the results of five methods,
and the average user ranking is used as a preference metric.

\noindent \textbf{Datasets.} 
We constructed a video test dataset consisting of 73 videos.
The majority of these videos are selected from the DAVIS~\citep{pont20172017} public dataset, 
which contains a diverse collection of semantically rich videos with pronounced motion.
Additionally, some videos are collected from Pixabay~\cite{pixabay}, 
featuring high-quality videos with significant motion.
All quantitative metrics are evaluated on our collected dataset.
For each video, two lighting prompts are applied, and three lighting directions are randomly chosen. 

\noindent \textbf{Implementation details.} Unless otherwise specified,
the default models employed for image relighting and 
VDM in the subsequent experiments are IC-Light~\citep{zhang2025scaling} 
and AnimateDiff~\citep{guo2023animatediff} Motion-Adapter-v3, respectively.
In the IC-Light model, the lighting conditions $c$ for image relighting are derived from two components:
First, a text prompt that describes the characteristics of the light source (e.g., neon light, sunshine, etc.).
Second, a lighting map is utilized to represent the light intensity across the scene.
This lighting map is then encoded by a VAE and serves as the initial latent for the denoising process.
During the inference stage, the source video is added with 50\% noise. 
Subsequently, the VDM employs a denoising process with $T_m = 25$ steps to 
progressively fuse the relight target.
For the parameters in the pipeline,
$\gamma=0.5$ in the CLA module is used to balance the original attention feature
and the cross-frame averaged feature.
In the PLF strategy, the fusion weight $\lambda_t$ decreases as denoising progresses,
and we set $\lambda_t = 1 - t/T_m$. 
\vspace{-0.5em}
\subsection{Qualitative Results}
\label{sec:QR}
As depicted in Fig.~\ref{fig:comparison},
the frame-by-frame IC-Light method ensures high single-frame quality.
However, the lack of consistency design and VDM temporal priors leads to
significant flickering of the light source and overall appearance. 
By introducing VDM priors, IC-Light + SDEdit-0.2 maintains content consistent with the source video, 
but still exhibits noticeable relight appearance jitter. 
IC-Light + SDEdit-0.6 further enhances temporal smoothness, 
yet object identity shifts occur.  
AnyV2V transfers the appearance of the first relight
frame to subsequent frames, 
but this pixel-level migration method, 
lacking perception of the given light source,
results in unreasonable illumination changes.
In contrast, Light-A-Video achieves high-quality video relighting, 
demonstrating strong temporal consistency and high fidelity to the light source.

\subsection{Video Relighting with Background Generation}
\label{sec:APP}
\begin{figure}[htp]
\centering
\includegraphics[width=\linewidth]{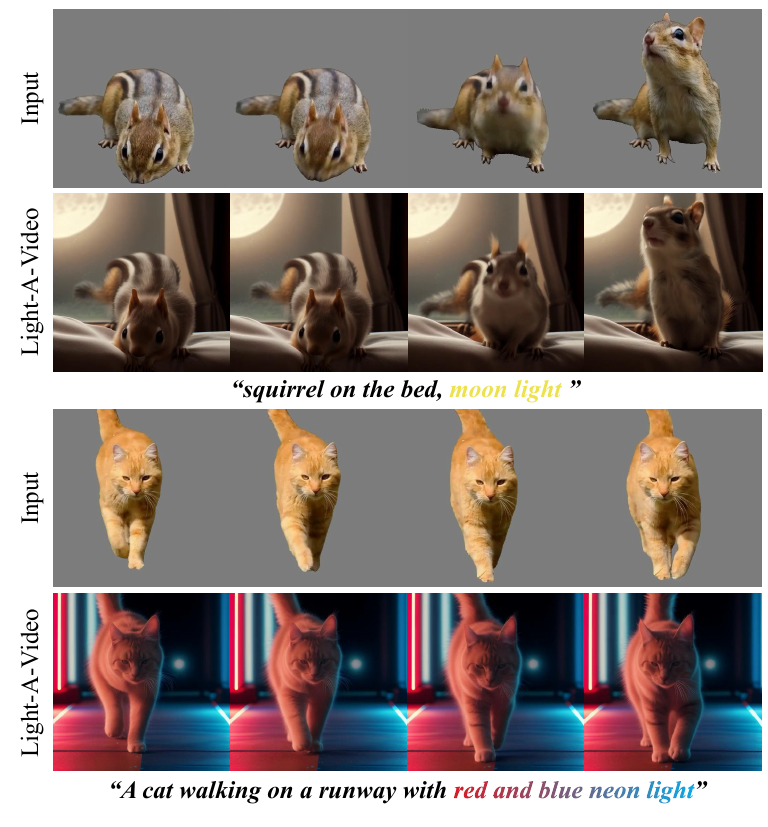}
\vspace{-2.5em}
\caption{{\bf Text-conditioned video illumination modifying with background generation}. 
Given a video foreground sequence and a text description of the target illumination, our method synthesizes suitable backgrounds and harmonious illumination.
}
\label{fig:inpaint}
\vspace{-1.5em}
\end{figure}

\vspace{-0.5em}
As depicted in Fig.~\ref{fig:inpaint}, Light-A-Video can accept a video foreground sequence and
a user-provided text prompt as input, generating a corresponding video background 
and illumination that aligns with the prompt descriptions.
Specifically, the input foreground sequence is initially processed
with IC-Light for frame-by-frame relighting,
while the background is entirely noised to serve as the initialization latent for the VDM.
From step $T$ to $T_m$, the background is progressively denoised,
leveraging the VDM's inpainting capability to generate the background.
Subsequently, from step $T_m$ to 0, the CLA module and PLF strategy are employed to achieve a temporally consistent  
relighting appearance of the video. 
These results illustrate that our pipeline
can produce high-quality video relighting results with consistent background generation. \footnote{More examples and ablation experiments are provided in the supplementary material.}


\subsection{Quantitative Evaluation}
\label{sec:QE}
The quantitative comparison of our method with various baselines is presented in Tab.~\ref{tab:quant}.
Given the addition of only 20\% noise, IC-Light + SDEdit-0.2 exhibits performance in video relighting 
that is nearly identical to IC-Light, resulting in significant temporal flickering in both methods. 
IC-Light + SDEdit-0.6 provides enhanced temporal consistency but suffers from object identity shifts due to the introduction of excessive noise.
For AnyV2V, the appearance of the first frame aligns well with the IC-Light results.
However, its inability to perceive the light source, 
combined with inherent quality degradation in subsequent frames, 
leads to a low motion preservation score. 
In contrast, Light-A-Video achieves a low FID score while maintaining high temporal consistency,
demonstrating its effectiveness in both relighted image quality and temporal stability.

\begin{figure}[htp]
\centering
\includegraphics[width=\linewidth]{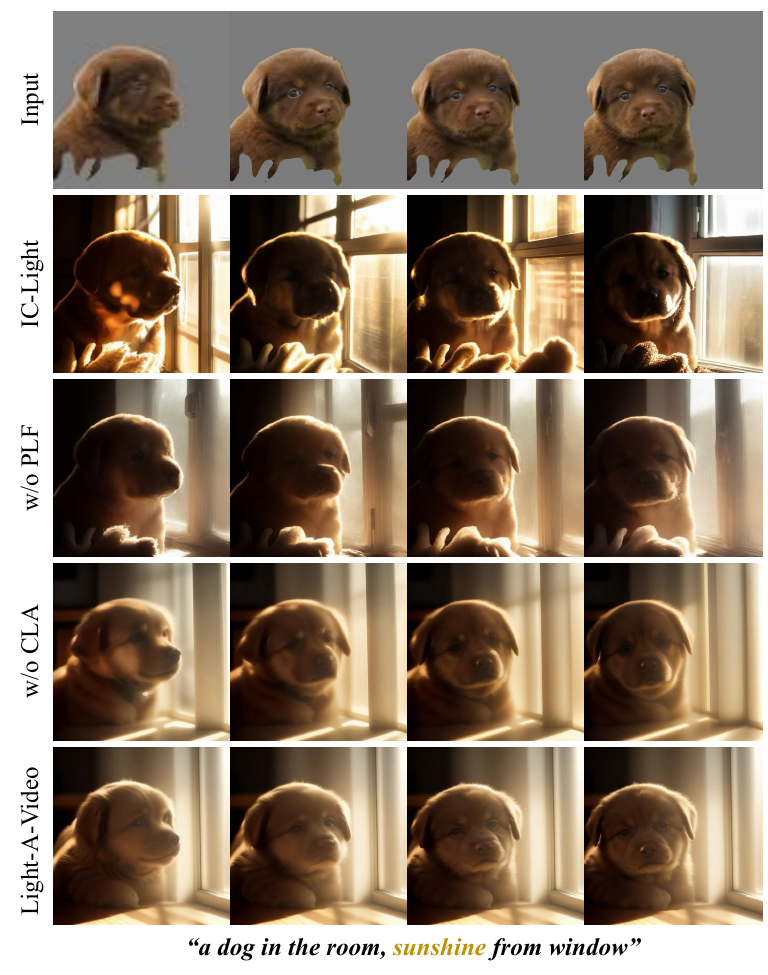}
\vspace{-2.5em}
\caption{{\bf Ablation Study}. 
Results of video relighting with the CLA module or the PLF strategy removed.
}
\label{fig:ablation}
\vspace{-1.5em}
\end{figure}

\vspace{-0.5em}
\subsection{Ablation Study}
\label{sec:abla}
An ablation study is conducted to assess the importance of our CLA and PLF modules.
As illustrated in Fig.~\ref{fig:ablation}, for the video relighting task involving background generation,
frame-by-frame IC-Light provides high-quality single-frame relighting but lacks control over temporal consistency.
This results in inconsistencies in lighting sources and relighted appearances across frames.
The CLA module enables cross-frame information exchange, which stabilizes the generation of background lighting sources.
Additionally, by introducing VDM motion priors and employing PLF's strategy for progressive fusion of relight targets 
into the original denoising target,
Light-A-Video ensures temporally smooth relighting.
The overall video quality is also significantly improved with the aid of VDM priors.

\vspace{-0.5em}
\subsection{Limitation and Future Work}
\label{sec:limitation}
Despite the impressive results achieved by our training-free method,
its performance is inherently constrained by the capabilities of the underlying image-relighting model and the VDM. 
While Light-A-Video demonstrates remarkable proficiency in ensuring stable lighting and temporal consistency, 
the CLA module, which is designed to stabilize background lighting,
exhibits limitations when it comes to modeling dynamic lighting changes.
To address this limitation, future work will focus on
developing novel methods that can more effectively handle dynamic lighting conditions. 

%% file: tabs/tab_1.tex
\begin{table*}
\footnotesize
\centering
\begin{tabular}{lcccccc}
\toprule[1.5pt]
\multirow{2}{*}{Evaluation Metric} & \multicolumn{1}{c}{(a) Relighted Image Quality} & \multicolumn{2}{c}{(b) Temporal Consistency} & \multicolumn{3}{c}{(c) User Preference} \\ 
\cmidrule(r){2-2} \cmidrule(r){3-4} \cmidrule(r){5-7}
   & FID Score ($\downarrow$) & CLIP Score ($\uparrow$)  & Motion Preservation ($\downarrow$) & LPA ($\uparrow$)  & VS ($\uparrow$) & IP ($\uparrow$) \\ \midrule[0.5pt]
IC-Light~\citep{zhang2025scaling}    & /  & 0.9040 &  5.969 & \underline{3.160}  & 2.128  & \underline{3.068} \\
IC-Light + SDEdit-0.2 & \textbf{13.79}
  & 0.9199 & \underline{5.959} & 2.850 & 2.752 & 3.014 \\
IC-Light + SDEdit-0.6 & 62.61
  & \underline{0.9483} & 7.544 & 2.472 & \underline{3.318} & 2.488 \\
IC-Light + AnyV2V~\citep{ku2024anyv2v} & 32.73
  & 0.9436 & 8.854  & 2.766 & 3.300 & 2.774 \\
\textbf{Light-A-Video (Ours)} & \underline{29.63}
    & \textbf{0.9667} & \textbf{1.833} & \textbf{3.752}  & \textbf{3.502}  & \textbf{3.656} \\ 
 \bottomrule[1.5pt]
\end{tabular}
\centering
\vspace{-1em}
\caption{\textbf{Quantitative comparison of baseline methods.} We achieves better results in relighted image quality and temporal stability. }
\vspace{-2em}
\label{tab:quant}
\end{table*}

%% file: sec/6_conclusion.tex
\vspace{-0.5em}
\section{Conclusion}
\label{sec:conclusion}
\vspace{-0.5em}
In summary, this paper introduces Light-A-Video, 
a training-free method that utilizes state-of-the-art image relighting models 
to achieve temporally consistent video relighting.
By incorporating the Consistent Light Attention (CLA) module to stabilize lighting source generation and 
employing the Progressive Light Fusion (PLF) strategy for smooth appearance transitions, 
Light-A-Video significantly enhances the temporal coherence of relighted videos 
while preserving the high-quality relighting of individual frames.

%% file: sec/supp.tex
\clearpage
\setcounter{page}{1}
\maketitlesupplementary
\appendix
\begin{figure}[htbp]
    \centering
    \includegraphics[width=0.9\columnwidth]{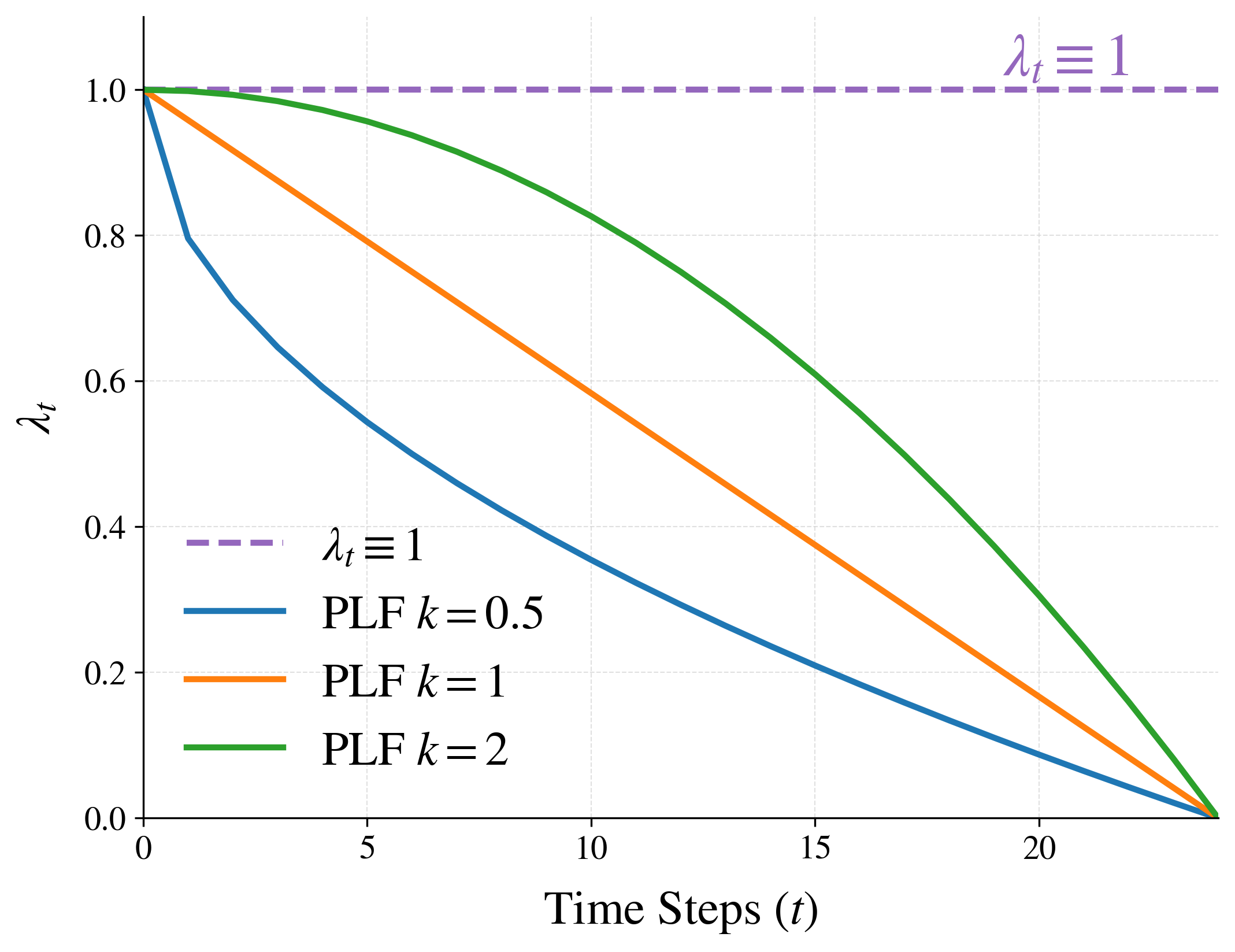}
    \caption{\textbf{Evolution of $\lambda_t$ over time steps $t$ for different PLF strategies.}
     $\lambda_t$ determines the proportion of the relight target mixed into the fusion target.
    }
    \label{fig:lambda}
    \vspace{-1em}
\end{figure}
\section{Comprehensive Ablation Studies}
In this section, we conduct comprehensive ablation studies to explore the effects of 
the hyper-parameter $\gamma$ of the Consistent Light Attention \textbf{(CLA)} 
and various Progressive Light Fusioin \textbf{(PLF)} strategies on the quality of relighted video generation.
Specifically, the values of $\gamma$ are uniformly sampled within the range of $[0,1]$, where
a larger $\gamma$ indicates a higher proportion of the cross-frame averaged feature in the CLA.
Notably, when $\gamma=0$, it corresponds to the vanilla IC-Light with standard self-attention.
For the PLF strategy, the parameter $\lambda_t$ determines
the proportion of the relight target mixed into the fusion target at each step.
Several different PLF strategies are also proposed, with $\lambda_t$ defined as:
\begin{equation}
\lambda_t = 1 - \left(\frac{t}{T_m}\right)^k
\end{equation}
Here, $T_m=25$ denotes the total number of noise-adding steps for the source video,
and different values of $k$ indicate different rates of decay for $\lambda_t$ over time.
$\lambda_t \equiv 1$ means directly replacing the fusion target with the relight target for all steps.
Fig.~\ref{fig:lambda} illustrates the curves of $\lambda_t$ as it varies with time step $t$.

\begin{figure*}[htbp]
    \centering
    \includegraphics[width=\linewidth]{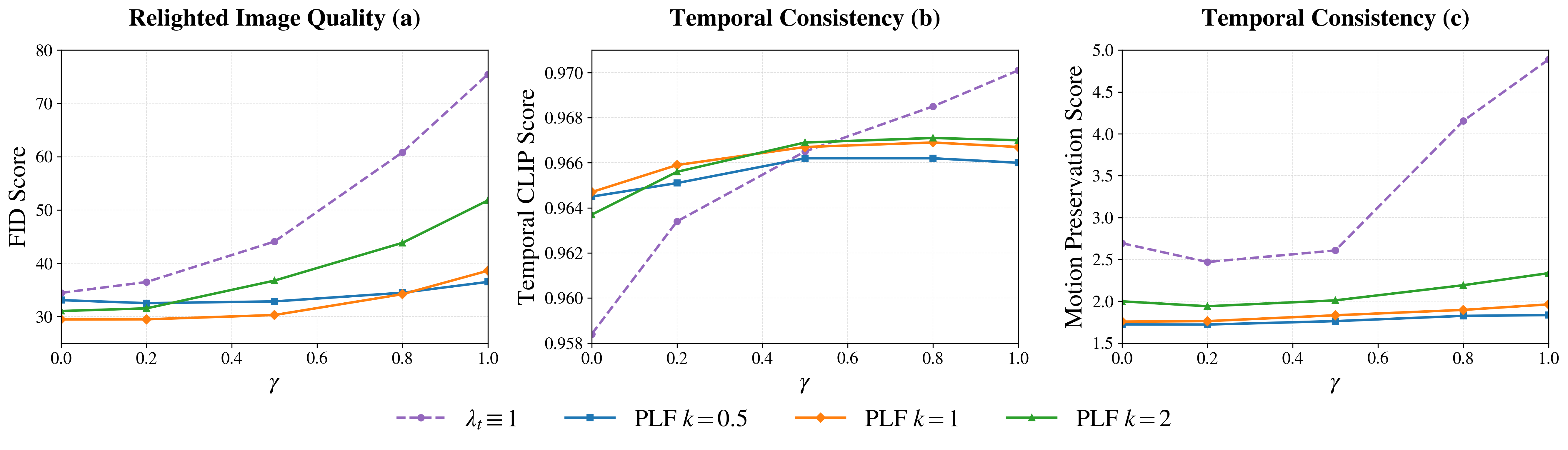}
    \caption{\textbf{The relative effectiveness of different PLF strategy on Light-A-Video performance.}
    (a) FID scores, (b) Temporal CLIP scores, and (c) Motion Preservation scores are shown for four strategies:
    PLF with constant $\lambda$ ($\lambda_t \equiv 1$), and PLF with $k=0.5, 1, 2$.
    Lower FID/Motion Preservation scores and higher Temporal Clip scores indicate better performance.
    }
    \label{fig:analysis}
\end{figure*}

A quantitative comparison of various settings is provided in Fig.~\ref{fig:analysis}, where the
three evaluation metrics (FID, Temporal Clip score, and Motion Preservation score) introduced in the main text are employed to
evaluate the per-frame image quality and temporal consistency of the relighted video generated by our Light-A-Video method.
Specifically, Fig.~\ref{fig:analysis} (a) depicts the variation of the FID score with different values of the trade-off parameter $\gamma$.
An excessively large $\gamma$ results in a significant degradation of the overall relighting image quality.
This is attributed to the overemphasis on the cross-frame averaged feature in the CLA module, 
which leads to temporal over-smoothing and diminishes the lighting specificity, 
thereby negatively impacting the relighting effect.
However, when $\gamma$ is chosen appropriately (between 0.2 and 0.5), the FID score remains stable and 
can even be enhanced, especially when employing PLF strategies with $k=1$ or $k=0.5$.

The temporal consistency evaluation, as depicted in Fig.~\ref{fig:analysis} (b), 
demonstrates a steady increase in the Temporal Clip score with the rise of the parameter $\gamma$.
This trend underscores the remarkable efficacy of the CLA module in augmenting the temporal consistency of the relighted video.
These results reflect that the CLA module is highly effective in enhancing the temporal consistency of the relighted video.
In a parallel vein, the Motion Preservation score serves as an indicator of motion consistency with the source video. 
Specifically, when the value of $\gamma$ is selected within the range of $0.2$ to $0.5$,
the relighted video can achieve a high degree of motion consistency with the original video.

It is worth noting that, as evidenced by the three figures, employing a constant $\lambda_t \equiv 1$ 
significantly underperforms the method of progressively decreasing $\lambda_t$ in PLF, both in terms of 
relight image quality and temporal consistency. Although a constant $\lambda_t$ yields a higher Temporal Clip score when
$\gamma > 0.5$, the overall motion deviates substantially from the source video, resulting in an unacceptable motion preservation effect.
These results effectively demonstrate the efficacy of our PLF strategy.
The explanation for this observation is twofold:
\begin{itemize}
\item Compared to a dynamically mixed target, a constant target with rich additional illumination information 
in the denoising process is more likely to deviate from the sampling trajectory of the Video Diffusion Model (VDM). 
When this deviation exceeds the refinement capability of the VDM, it perturbs the motion priors, consequently leading to visible temporal jitter.

\item Repeatedly injecting constant relight appearance across multiple iterations is analogous to cyclically relighting the same image using the image relight model. 
This process causes the input distribution to progressively diverge from the training distribution of the image relight model, ultimately degrading the quality of the relighted images.
\end{itemize}

\section{Additional Results}
\begin{figure*}[htbp]
    \centering
    \includegraphics[width=\linewidth]{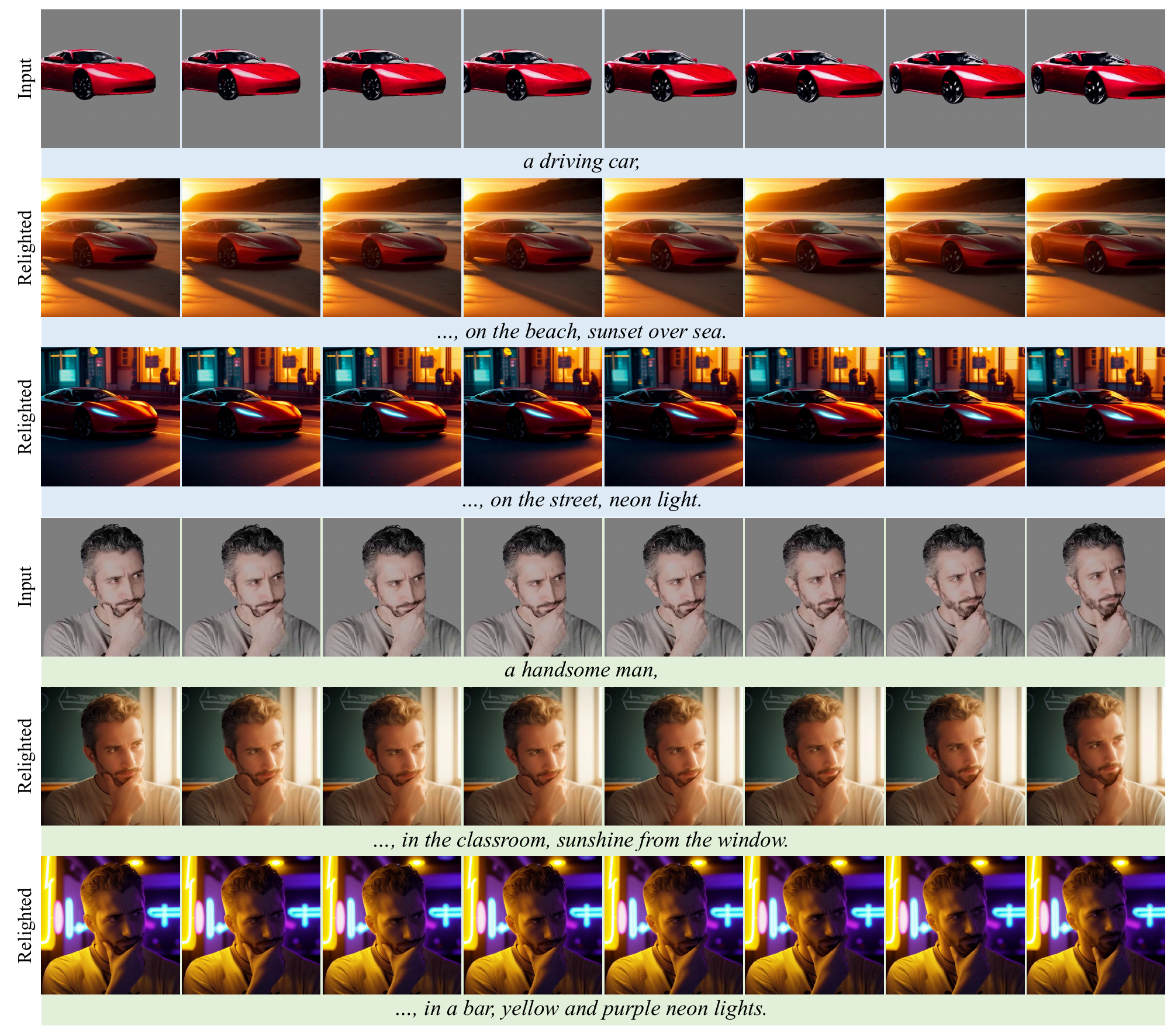}
    \caption{\textbf{More results of Light-A-Video in foreground sequences relighting with background generation.}}
    \label{fig:fig1}
\end{figure*}

\begin{figure*}[htbp]
    \centering
    \includegraphics[width=\linewidth]{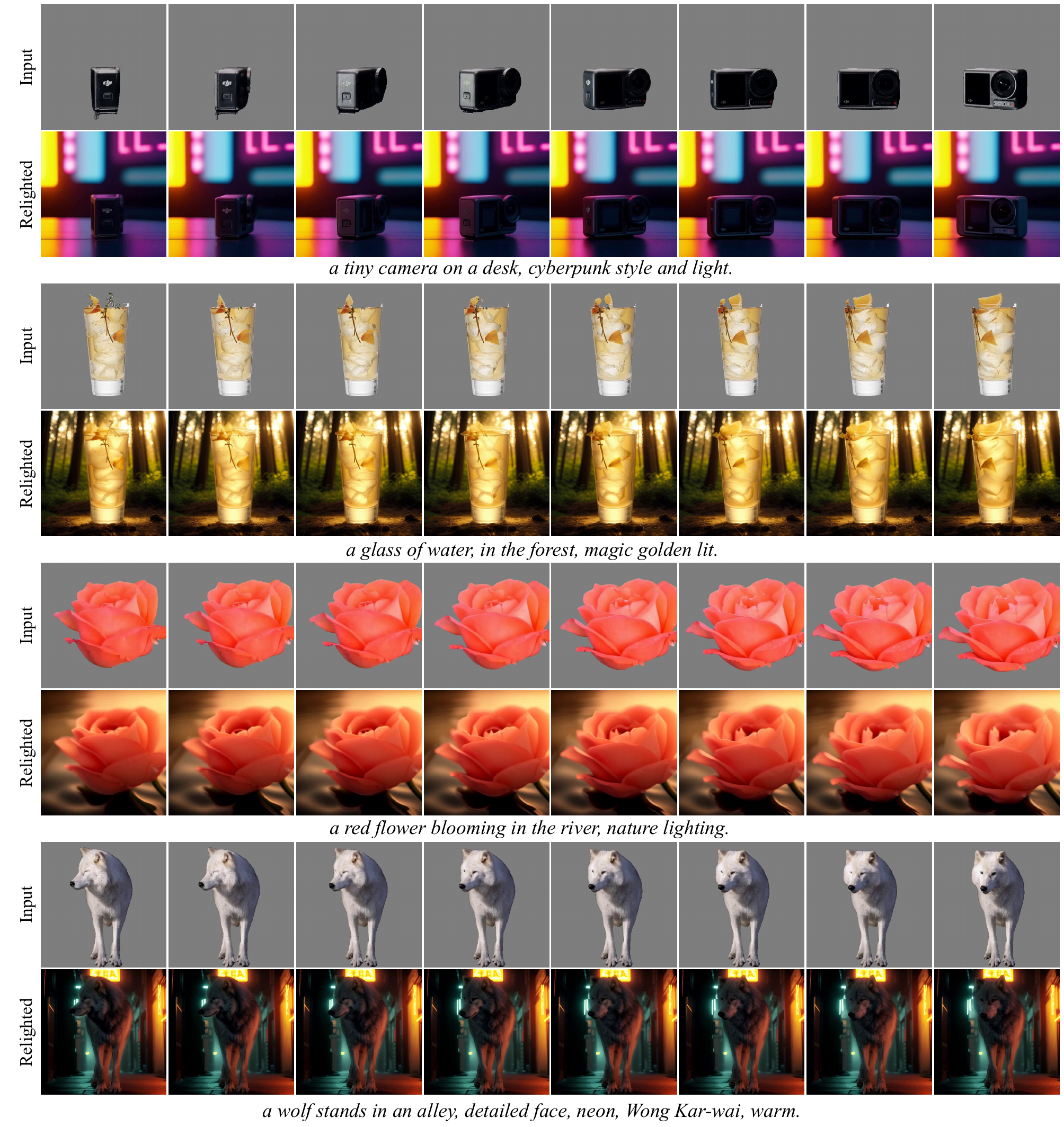}
    \caption{\textbf{More results of Light-A-Video in foreground sequences relighting with background generation.}}
    \label{fig:fig2}
\end{figure*}

\begin{figure*}[htbp]
    \centering
    \includegraphics[width=\linewidth]{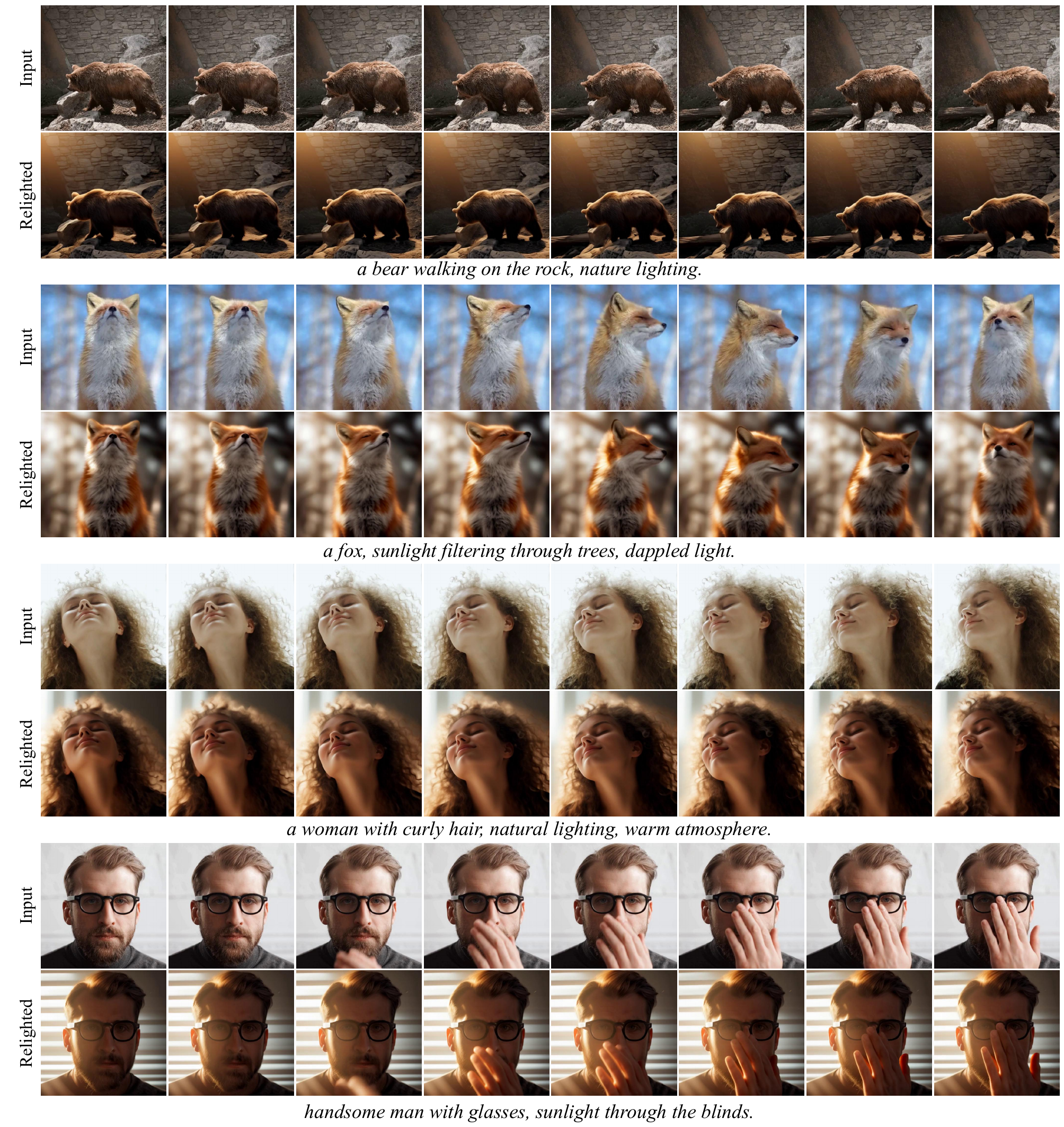}
    \caption{\textbf{More results of Light-A-Video in video sequences relighting.}}
    \label{fig:fig3}
\end{figure*}

\begin{figure*}[htbp]
    \centering
    \includegraphics[width=\linewidth]{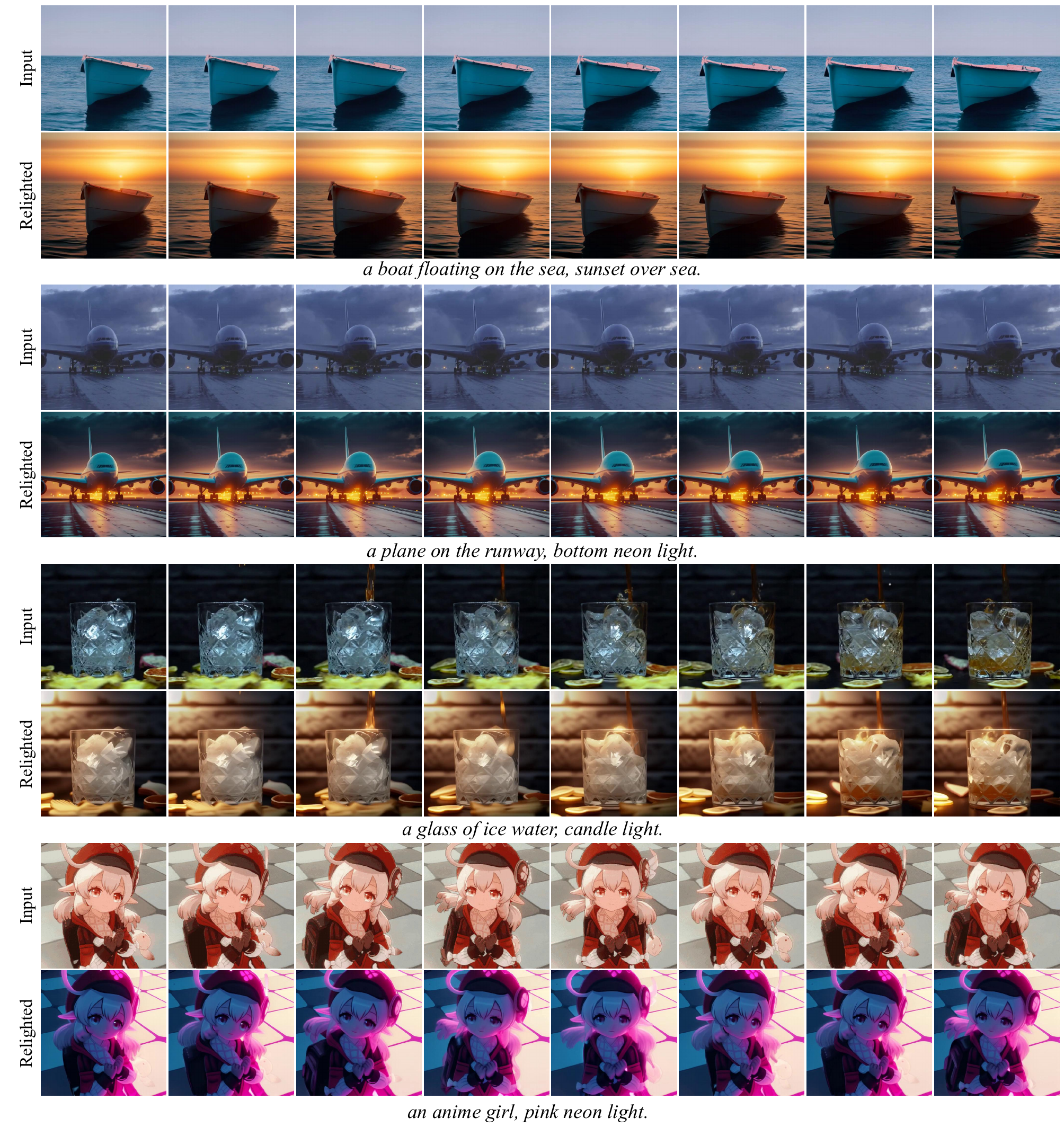}
    \caption{\textbf{More results of Light-A-Video in video sequences relighting.}}
    \label{fig:fig4}
\end{figure*}

\begin{figure*}[htbp]
    \centering
    \includegraphics[width=\linewidth]{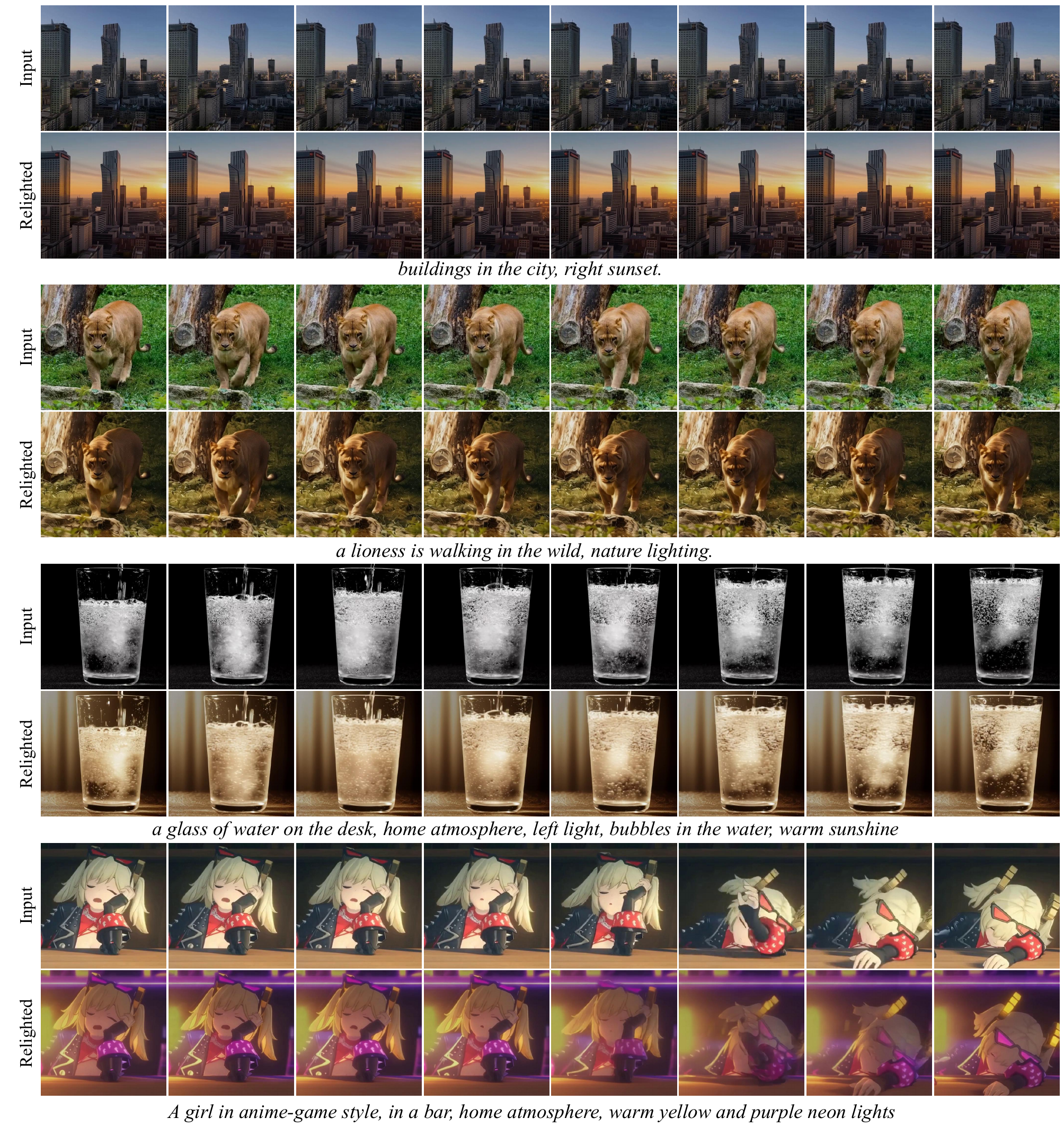}
    \caption{\textbf{More results of Light-A-Video in video sequences relighting on CogVideoX.}}
    \label{fig:fig5}
\end{figure*}

In this section, we present additional qualitative results.
In Fig.~[\ref{fig:fig1}-\ref{fig:fig2}], we show examples of foreground sequences relighting with background generation on AnimateDiff.
In Fig.~[\ref{fig:fig3}-\ref{fig:fig4}], we showcase the application of Light-A-Video directly to the video relighting task. 
And finally, as illustrated in Fig.~\ref{fig:fig5}, we present the video relighting results on DiT-based video models, such as CogVideoX.

